\documentclass[utf8]{FrontiersinHarvard} 
\usepackage{url,lineno,microtype,subcaption,nicefrac}
\usepackage[onehalfspacing]{setspace}
\usepackage[colorlinks=true,linkcolor=blue,citecolor=blue]{hyperref}
\newcommand{\modified}[1]{\textcolor{black}{#1}}
\setcitestyle{numbers}

\def\keyFont{\fontsize{8}{11}\helveticabold }
\def\firstAuthorLast{Sommer {et~al.}} 
\def\Authors{Philipp Sommer\,$^{1,3}$, Yixing Huang\,$^{1,2,*}$, Christoph Bert\,$^{1,2}$, Andreas Maier\,$^{3}$, Manuel Schmidt\,$^{4}$, Arnd D\"orfler\,$^{4}$, Rainer Fietkau\,$^{1,2}$ and Florian Putz\,$^{1,2,*}$}
\def\Address{$^{1}$Department of Radiation Oncology, University Hospital Erlangen, Friedrich-Alexander-Universität Erlangen-Nürnberg, Erlangen, Germany\\
$^{2}$Comprehensive Cancer Center Erlangen-EMN (CCC ER-EMN), 91054 Erlangen, Germany \\
$^{3}$ Pattern Recognition Lab, Friedrich-Alexander-Universität Erlangen-Nürnberg, Erlangen, Germany\\
$^{4}$ Department of Neuroradiology, University Hospital Erlangen, Friedrich-Alexander-Universität Erlangen-Nürnberg, Erlangen, Germany 
}
\def\corrAuthor{F. Putz \textit{orcid.org/0000-0003-3966-2872}; Y. Huang \textit{orcid.org/0000-0003-2627-3077}}

\def\corrEmail{florian.putz@uk-erlangen.de; yixing.huang@uk-erlangen.de}

\begin{document}

\onecolumn
\firstpage{1}

\title[Risk Classification of Brain Metastases]{Risk Classification of Brain Metastases via Radiomics, Delta-Radiomics and Machine Learning}

\author[\firstAuthorLast ]{\Authors} 
\address{} 
\correspondance{} 

\extraAuth{}

\maketitle

\begin{abstract}


Stereotactic radiotherapy (SRT) is one of the most important treatment for patients with brain metastases (BM). Conventionally, following SRT patients are monitored by serial imaging and receive salvage treatments in case of significant tumor growth. We hypothesized that using radiomics and machine learning (ML), metastases at high risk for subsequent progression could be identified during follow-up prior to the onset of significant tumor growth, enabling personalized follow-up intervals and early selection for salvage treatment. All experiments are performed on a dataset from clinical routine of the Radiation Oncology department of the University Hospital Erlangen (UKER). The classification is realized via the maximum-relevance minimal-redundancy (MRMR) technique and support vector machines (SVM). The pipeline leads to a classification with a mean area under the curve (AUC) score of $0.83$ in internal cross-validation and allows a division of the cohort into two subcohorts that differ significantly in their median time to progression (low-risk metastasis (LRM): $17.3$ months, high-risk metastasis (HRM): $9.6$ months, $p < 0.01$). The classification performance is especially enhanced by the analysis of medical images from different points in time (AUC $0.53$ $\rightarrow$ AUC $0.74$). The results indicate that risk stratification of BM based on radiomics and machine learning during post-SRT follow-up is possible with good accuracy and should be further pursued to personalize and improve post-SRT follow-up.

\tiny
 \keyFont{ \section{Keywords:} Radiomics, BM, machine learning, Stereotactic radiosurgery, Tumor progression, follow-up care} 
\end{abstract}

\section{Introduction}
\modified{Brain metastases (BM) appear frequently in systematic cancer disease with an incidence rate up to 40\%, depending on the type and initial stage of primary tumor \cite{tabouret2012recent}.} The treatment of BM is highly demanding but challenging: Surgeries are oftentimes too risky or infeasible; chemotherapy fails because of the blood-brain barrier; whole brain radiotheraphy (WBRT) leads to severe side effects. \modified{Nowadays, stereotactic radiosurgery (SRS) with highly focused radiation beams has become the main stream for BM treatment due to its high efficacy and minimal side effects \cite{Lowery.2017,Wannenmacher2013}. Depending on treatment responses, local control of BM is pronounced while progression and radionecrosis are also frequently observed in patients after treatment \cite{putz2020fsrt,oft2021volumetric}, along with other treatment effects like inflammation and vascular injury. Hence, treatment prognosis and follow-up care play a crucial role in patient survival and quality of life \cite{le2017eano}. 
}

\modified{Machine learning (ML) gains more and more influence in modern radiation oncology \cite{huang2022deep,weissmann2023deep}. For BM treatment, ML has been applied to various stages from detection and segmentation \cite{xue2020deep,huang2022deep} to treatment response \cite{kawahara2021predicting,jaberipour2021priori} and survival prediction \cite{huang2019mining,bice2020deep}. BM treatment require advanced nuclear imaging such as planning computed tomography (CT) and multi-sequence magnetic resonance imaging (MRI) \cite{kaufmann2020consensus}. Such imaging data not only provides essential information for BM treatment but also contains potential correlation with patient survival after treatment. However, imaging data is high-dimensional and thus infeasible for a straightforward interpretation of its underlying information. Quantifying information from medical images using ML} allows to turn the power of mathematics and statistics into motion towards personalized and highly precise cancer treatments -- including follow-up care. \modified{In particular, the techniques of radiomics receive a lot of interest recently \cite{Gillies.2016}.}

\modified{Radiomics extracts a large number of features from medical images to uncover tumor characteristics \cite{kocher2020applications,lohmann2020pet}. The features can be extracted via known operators, which offers high interpretability of the handcrafted features. Alternatively, features can be directly learned by a neural network, where high-level effective but abstract features are extracted. Radiomics based on such handcrafted or learned features have been investigated in various topics related to BM, where encouraging results have been reported. For example, Stefano et al. have conducted a preliminary radiomics study of BM on PET data, where 3 and 8 features among ectracted 108 features are indicated to have valuable association with patient outcome.
Ahn et al. \cite{ahn2020contrast} suggested that radiomic features extracted from contrast enhanced T1-weighted images can predict the EGFR mutation status of lung cancer BM.
Artzi et al. \cite{artzi2019differentiation} have applied radiomics analysis to differentiate glioblastom, BM and subtypes. 
Kniep et al. \cite{kniep2019radiomics} investigated the feasibility of using radiomics to distinguish brain metastatic tumor type.
Mouraviev et al. \cite{mouraviev2020use} applied a random forest classifier with radiomic features for the prediction of BM local control after SRS. 
Peng et al. \cite{peng2018distinguishing} concluded that ML and radiomics can distinguish true progression from radionecrosis in BM after SRS.
For deep learning radiomics, Bae et al. \cite{bae2020robust} and Liu et al. \cite{liu2021handcrafted} both demonstrated that deep learning radiomics can distinguish glioblastoma from single BM with good generalizability.
Yang et al. \cite{yang2022deep} proposed a ensemble classifier combining deep learning with radiomics, which can reduce flase positve segmentation of BM.
A comprehensive review of radiomics in BM can be found in \cite{kocher2020applications,lohmann2020pet}.
}

\modified{In this work, radiomics and ML are applied for risk classification of BM towards personalized follow-up care. The current practice of} follow-up care is a new scan every three months after SRS -- roughly 100 days \cite{Wannenmacher2013}. Despite including relatively frequent imaging, the follow-up scheme might not be suited for patients with BM that show an extremely high risk of progression, in the following referred to as high-risk metastasis (HRM): Tumor growth might appear within this very 100 days and will therefore stay undetected until the date of imaging. To enlighten this blind spot, this ML based technique is presented. \modified{The contributions of this work mainly lie in the following aspects:}

\begin{itemize}
    \item To the best of our knowledge, our work is the first study to apply radiomics for BM risk classification in a patient cohort containing both tumor volumetric growth and regression.
    
    \item The efficacy of delta-radiomic features extracted from a longitudinal dataset for BM risk classification is investigated for the first time.
\end{itemize}


\section{Materials and Methods} 

\subsection{Dataset} \label{sec:materials}

The data used in this study originates from a longitudinal dataset of BM patients treated with SRS at the University Hospital Erlangen between 2003 and 2015. Already introduced by \cite{Oft2021}, the data set includes 189 patients and 419 BM in total, captured in 292 CT volume images and 1394 MRI volume images. While the CT images all origin from one device, the MRI images were taken from 10 different devices (7 model types) and over 10 sequences in total. 

In opposite to prominent works \cite{Aerts.2014,VanGriethuysen2017,Gillies.2016} in the field of radiomics, this study was on purpose performed on this real world dataset. This fulfills two major tasks: Providing deeper insights to the data set itself and showing the capability of radiomics on unseen raw data.

\begin{figure} [h]
\begin{center}
\includegraphics[width=1\linewidth]{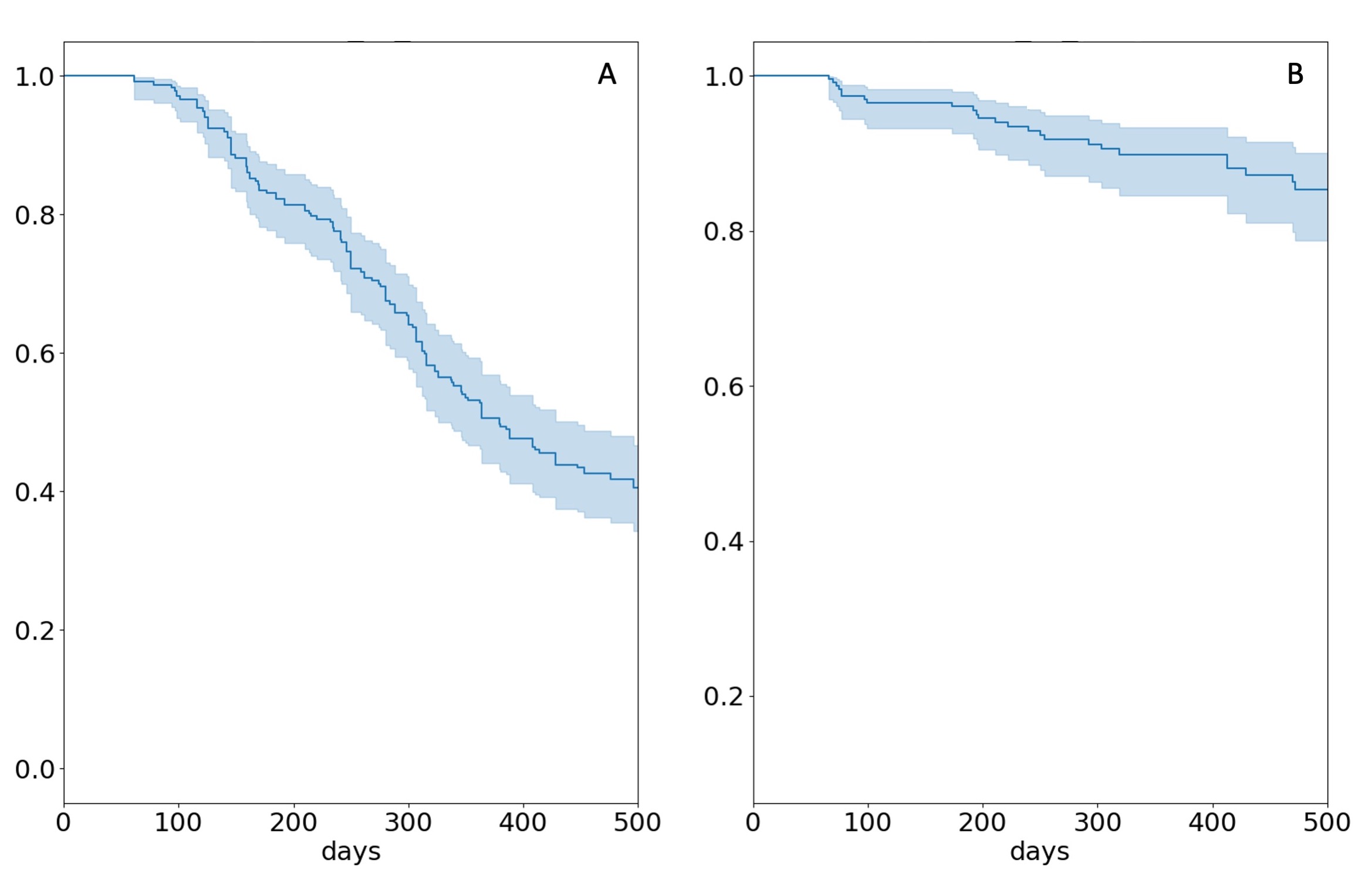}
\end{center}
\caption{KPM Plots of survival and freedom of progression over time. (A) Survival chances over time. (B) Risk of freedom of progression over time. Blue area indicates 95$\%$ confidence interval.}\label{fig:kpm}
\end{figure}

At studies end in 2015, 84$\%$ (159/189) patients had died, while 16$\%$ (30/189) were still alive or lost to follow-up. In 419 BM there were a total of 93 progression events. The chances of survival and local control are displayed in the KPM-Plot in \autoref{fig:kpm}.

\newpage
\subsection{Classification Pipeline}

The methods provided in this study follow well-known schemes in the field of radiomics, yet the focus on the changes in the medical images over time adds a special characteristic. These changes are manifested as delta-radiomics (See \autoref{eqn:delta}) and furthermore the separate use of imaging data from different points in time. 

\subsubsection{Data preparation} \label{data_prep}

For the classification, all imaging data and personal information described in Subsection \ref{sec:materials} that was accessible was used. The interval used for the classification between LRM and HRM starts from the point of the last imaging. It has to be noted that this interval is a continuous value, but only sampled at discrete points in time. That adds uncertainties and bias to the data: Changes in the metastases can only be observed at the time of imaging, not at the actual point in time, when the change occurred. 

From the originally 419 BM, a total of 292 BM was examined, as for the remaining 127 BM, no dose and Planning CT information was accessible. In total, 932 MRI images and 130 Planning CT images were analyzed: Each MRI image was labeled as HRM or LRM according to whether there was an progression event within the next 100 days from the time of imaging. 40 BM were classified as HRM, 892 as LRM.

MRI is the leading imaging modality in this dataset. Although providing undeniable benefits in the planning of SRS, MRI is a qualitative imaging technique and has to be treated with care \cite{Putz2020}. Therefore, the most possible harmonization between the images has to be ensured in order to produce meaningful and comparable metrics: All images were z- and white-stripe-normalized \cite{Shinohara2014} in order to minimize inter-device variability. Nevertheless, the variety in MRI devices adds differences to the images, as also discussed in Subsection \ref{sec:materials}. 

Wavelet filtering was applied to the medical images following other studies in the field of radiomics \cite{Gillies.2016}. In wavelet filtering, a combination of several high- and low-pass filters and small wave-like shaped forms (Wavelet) allows a filtering with beneficial properties as it is also known throughout the world of digital imaging \cite{Xizhi2008}. The creation of these images was realized using the \textit{pyWavelet} software package \cite{Lee2019a}.

\begin{figure} [h]
	\begin{center}
		\includegraphics[width=1\linewidth]{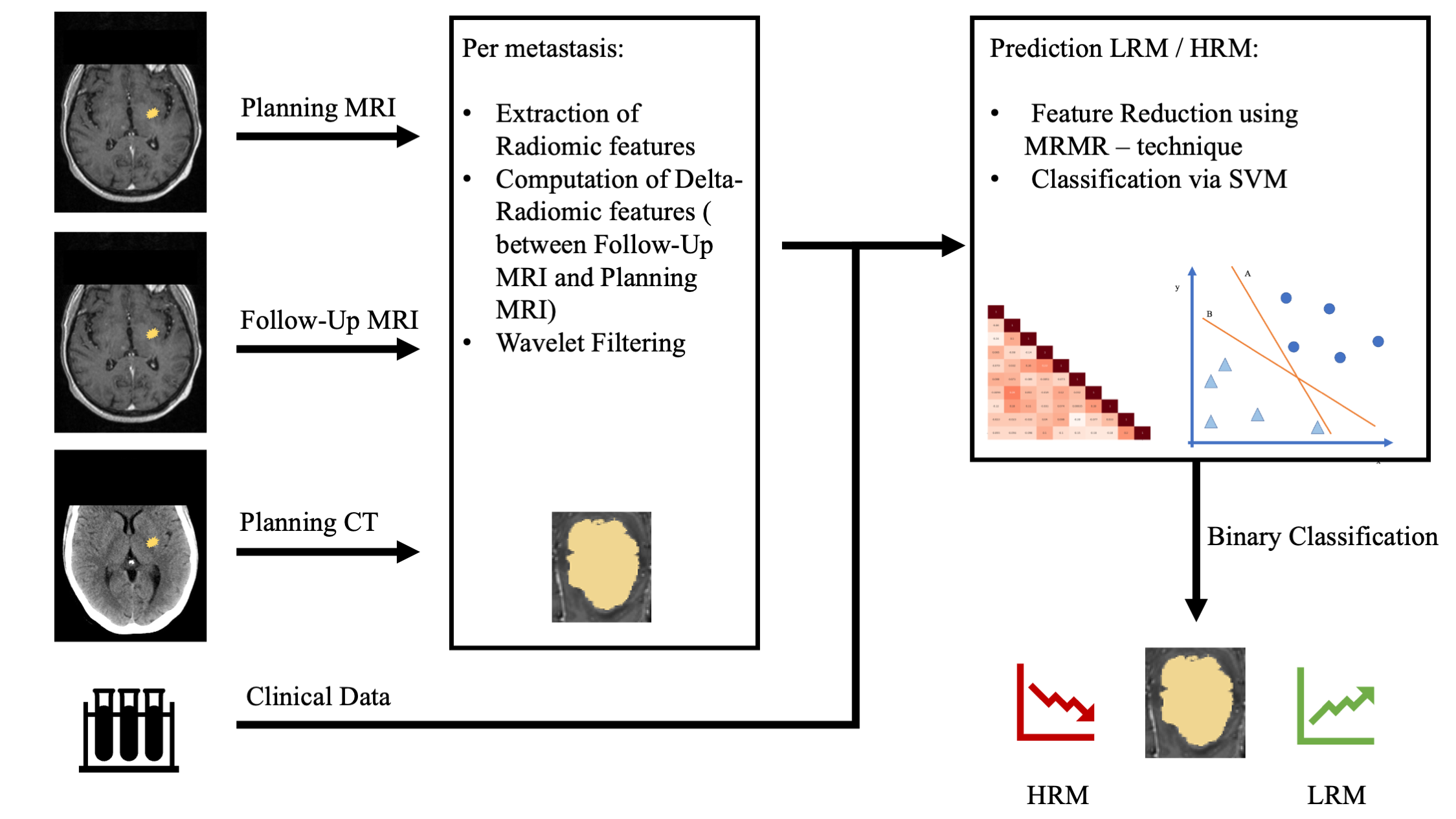}
	\end{center}
	\caption{Overview of the classification process. Features from Clinical Data and extracted radiomic and computed delta-radiomic features. Feature Reduction via MRMR-technique and binary classification of metastases into LRM and HRM using SVM. }
	\label{fig:ovw}
\end{figure}

In \autoref{fig:ovw}, the different steps within the classification pipeline are displayed to provide an overview of the executed steps. Each follow-up MRI is linked to its Planning MRI and Planning CT in order to follow the development of the metastasis. Further clinical data is added. The radiomic features are extracted from the medical images and the features from the clinical data (See Subsection \ref{feature extraction}) are filtered and reduced in the following (See Subsection \ref{feature selection}) and finally classified into HRM and LRM via an support vector machine (SVM) (See Subsection \ref{classification}).

\subsubsection{Feature extraction} \label{feature extraction}

\begin{figure} [h]
	\begin{center}
		\includegraphics[width=1\linewidth]{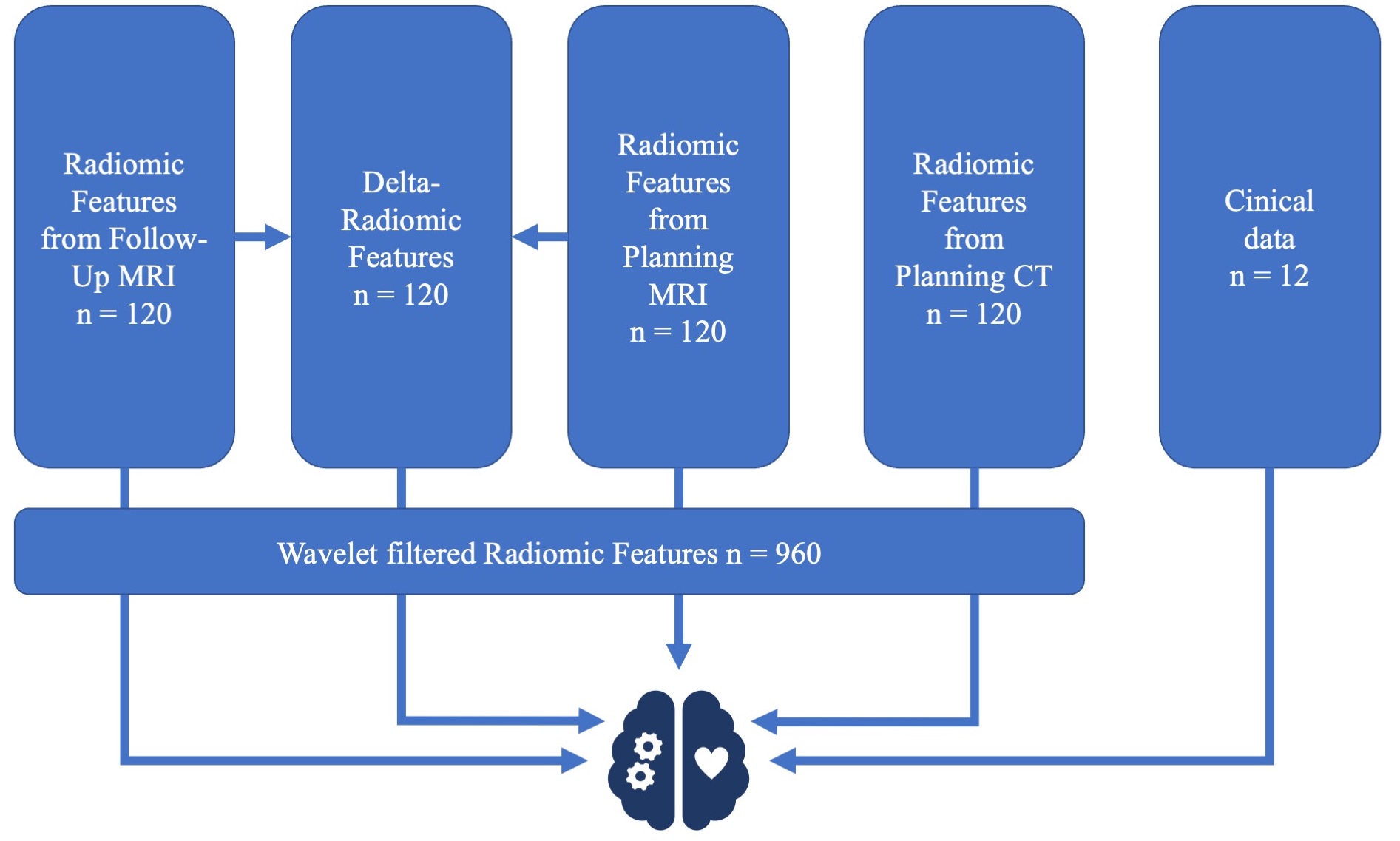}
	\end{center}
	\caption{Origin of features for risk classification of BM}\label{fig:overview_features}
\end{figure}

The extraction of radiomic features was realized via the \textit{python} software package \textit{pyradiomics}, a recent and broadly used tool for radiomic feature extraction \cite{VanGriethuysen2017}. From all images, including the wavelet filtered images, 120 features were extracted. They are divided into features describing the shape, first and higher order characteristics. For further information please see \cite{VanGriethuysen2017}. Adding 12 features from clinical data, the features sum up to over 4000 features in total. An overview is provided in \autoref{fig:overview_features}.

The delta-radiomic features were defined following the work of \cite{Fave2017}. The changes in the radiomic features were divided by the days passing, highlighting fast appearing changes. Let $F_{FU}$ be the radiomic feature from the follow-up MRI and $F_P$ the corresponding radiomic feature from the Planning MRI sequence. The days passed are denoted by $d$. The value for the Delta-radiomic feature $\Delta F$ is defined as:

\begin{equation} 
	\Delta F = \frac{F_{FU} - F_D}{d}\,.
	\label{eqn:delta}
\end{equation}

\subsubsection{Feature selection} \label{feature selection}

The feature selection was performed based on the principles described by \cite{HanchuanPeng2005} and is referred to as the MRMR-technique. Briefly, the algorithm neglects the features that show a low correlation to the label and a high correlation to one another. In other words, only features that correlate to the appearance of the label and add information, which is not yet provided by other features, are present after the feature selection. The selection was optimized to keep one feature per ten samples as suggested by \cite{Gillies.2016} to prevent overfitting and to keep the performance of the following classification step fast and robust.

\newpage
The correlation between the features, and between the label and the features, is assessed by the Pearson linear correlation coefficient $r_{X,Y}$ (\autoref{pearson coefficient}) \cite{Kirch2008}, which is defined as:

\begin{equation}
	r_{X,Y} = \frac{\sum_{i=1}^{n}(X_i - \overline{X})(Y_i - \overline{Y})}{\sqrt{\sum_{n}^{i=1}(X_i - \overline{X})^2}\sqrt{\sum_{i=1}^{n}(Y_i - \overline{Y})^2}}\,.
	\label{pearson coefficient}
\end{equation}

For two feature vectors $X,Y$ the value can vary between $+1$ and $-1$. $n$ denotes the number of samples. For high positive values, approaching $r_{X,Y}\rightarrow1$, a strong positive correlation is observed; for high negative values approaching$r_{X,Y}\rightarrow-1$, a strong negative correlation is observed. Small values for $r_{X,Y}\rightarrow0$ point to a very weak correlation. 

\subsubsection{Classification} \label{classification}

An SVM was chosen as the algorithm for the risk prediction of BM. Even less frequently used in literature \cite{VanGriethuysen2017,Zwanenburg.2019,Zwanenburg.2020}, the SVM was chosen mainly because of its relatively low sensitivity to class imbalance \cite{Suthaharan2016}. Also, its simplicity and fast convergence are well suited characteristics for this task. Further, the weighting of the classification's sensitivity versus the classification's specificity was performed according to clinical requirements, with the classification focusing on high sensitivity for detecting high-risk metastases allowing to sacrifice specificity therefore. Please refer to Section \ref{sec:discussion} for further detail.

\section{Results} \label{sec:results}

As already shown in Subsection \ref{data_prep}, the 932 samples were labeled as HRM and LRM according to their time to progression from the time of imaging. For each sample, over 4000 features were available, reduced to 93 features using the MRMR technique. Based on these features, the classification was performed.

\subsection{Classification performance}

\begin{figure} [h]
	\begin{center}
		\includegraphics[width=1\linewidth]{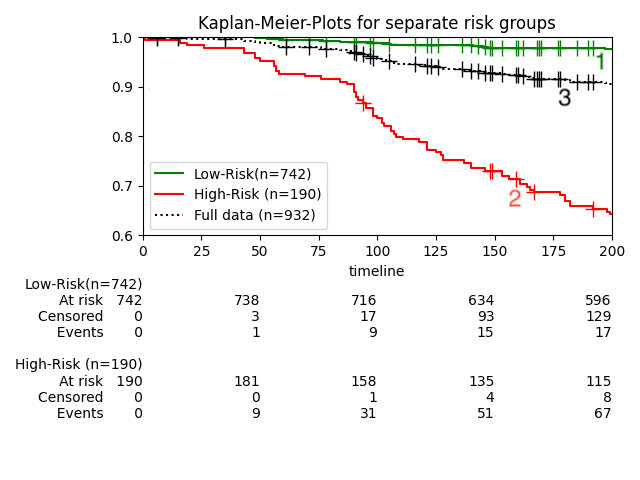}
	\end{center}
	\caption{Kaplan-Meier Plots of classified BM. Green line (1) indicating LRM, Red line (2) indicating HRM, black dashed line (3) showing complete cohort. Classified metastases differ significantly ($p < 0.01$).}
	\label{fig:kpm_split}
\end{figure}


The median time to progression in the HRM cohort is 9.6 months and 17.3 months for the LRM cohort, respectively (Significantly differing, p $<$ 0.01). The timeline of 200 days is displayed on the x-axis. By day 100, 40 metastases showed progression and 18 were censored. 31 of 40 progressing metastases were correctly identified as HRM (true positive), 9 were falsely classified as LRM (false negative). Excluding censoring, 716 of 874 metastases were correctly identified as LRM (true negative) and 158 mis-classified as HRM (false positive). The classification performance, here visible by the separation of the red (2) and green (3) line, improves steadily as time passes. 

The classification was performed with cross validation to ensure higher generalization and lower chance of running into overfitting \cite{Schaffer1993}. For each cross validation, the dataset is split into \nicefrac{2}{3} training data and \nicefrac{1}{3} test data. The procedure was repeated 100 times.

\subsection{Influence of feature sets} \label{sec:res_inf_sets}

Besides evaluating the performance of the classification system, the influence of different feature sets was analyzed. As depicted in \autoref{fig:overview_features}, the features originate from multiple sources, combining imaging at different points in time, wavelet filtered images and clinical data. The different data sets for the following experiments are Summarized in \autoref{tab:sets_risk}.

\begin{onehalfspacing}
	
	\begin{table} 
		\centering
		\begin{tabular}{l c c c c c c c}
			&Set 1				&Set 2				&Set 3				&Set 4				&Set 5					&Set 6 & Set 7\\
			\hline
			\hline
			Clinical data						& \checkmark & \checkmark & \checkmark & \checkmark & \checkmark & \checkmark & \checkmark \\
			
			Radiomic features follow-up MRI&  					& \checkmark &                       &                       &                    & \checkmark & \checkmark \\
			
			Delta-radiomic features	 				&  						&  					& \checkmark &                        &                & \checkmark& \checkmark  \\
			
			Radiomic features planning MRI				&  						&  					&  					& \checkmark &                     & \checkmark& \checkmark  \\
			
			Radiomic features planning CT					&  					&  						&  					&  						& \checkmark & \checkmark & \checkmark \\
			
			Wavelet filtered images								&  					&  						&  					&  						& 					 & 						& \checkmark  \\
			\hline
			\hline
			AUC score&$0.58$&$0.7$&$0.65$&$0.74$&$0.69$&$0.75$&$0.83$ \\
			\hline
			\hline
		\end{tabular}
		
		\caption{Overview of feature sets for the risk classification in BM and AUC-Score of the classification.}
		\label{tab:sets_risk}

		\vspace{36pt}
		\centering
		\begin{tabular}{clc}
			Rank & Feature name & $r_{F,L}$ \\
			\hline
			\hline
			\multicolumn{3}{l}{Clinical data} \\
			\hline
			1 & RPA class & $-0.2$ \\
			
			2 & EQD & $-0.18$  \\
			
			3 & Number of metastasis & $-0.09$  \\
			\hline
			\multicolumn{3}{l}{Radiomic features of follow-up MRI} \\
			\hline
			1 & follow-up-mr-original-shape-SurfaceVolumeRatio& $-0.15$ \\
			
			2 & follow-up-mr-original-shape-MajorAxisLength & $0.13$  \\
			
			3 & follow-up-mr-original-shape-Maximum2DDiameterRow & $0.13$ \\
			\hline
			\multicolumn{3}{l}{Delta-radiomic features} \\
			\hline
			1 & Delta-mr-original-shape-MinorAxisLength  & $0.08$ \\
			
			2 &Delta-mr-original-shape-SurfaceVolumeRatio& $-0.07$  \\
			
			3 & Delta-mr-original-shape-MajorAxisLength & $0.06$ \\
			\hline
			\multicolumn{3}{l}{Radiomic features of planning MRI} \\
			\hline
			1 & Plan-mr-original-gldm-LargeDependenceLowGrayLevelEmphasis  & $0.08$ \\
			
			2 & Plan-mr-original-gldm-SmallDependenceHighGrayLevelEmphasis & $-0.07$  \\
			
			3 &Plan-mr-original-gldm-SmallDependenceEmphasis  & $-0.06$ \\
			\hline
			\multicolumn{3}{l}{Radiomic features of planning CT} \\
			\hline
			1 & Plan-ct-original-shape-Sphericity   & $-0.08$ \\
			
			2 & Plan-ct-original-shape-Elongation & $-0.08$  \\
			
			3 & Plan-ct-original-glszm-SizeZoneNonUniformityNormalized & $-0.08$ \\
			\hline
			\multicolumn{3}{l}{Wavelet-filtered radiomic features} \\
			\hline
			1 & follow-up-mr-wavelet-LHL-firstorder-Range   & $0.2$ \\
			
			2 & follow-up-mr-wavelet-HHL-firstorder-Maximum & $0.2$  \\
			
			3 & follow-up-mr-wavelet-HHL-firstorder-Range & $0.19$ \\
			\hline
			\hline
			
		\end{tabular}
		\caption{Features with strongest correlation to label for each feature set.}
		\label{tab:risk_all_feat}
	\end{table}
	
\end{onehalfspacing} 

Solely using the features from clinical information, the area under curve (AUC) score reaches $0.58$. All added feature sets improve this score to values from $0.69$ to $0.74$. The combination of the different feature sets and the further use of wavelet filtering on the all imaging data allow a classification with an AUC score of $0.83$.

\subsection{Influence of single features}

The influence of single features is assessed by the correlation to the label value, comparable to the idea of the MRMR-technique. Doing so, for each feature set the most influencing features can be ordered and are displayed in \autoref{tab:risk_all_feat}. Highest absolute values for the correlations are observed for features from wavelet filtered images as well as features from the clinical information of a patient. 

\subsection{Influence of multiple time points}

Focusing on the delta-radiomic features and in general on the importance of an assessment of multiple time points, the features extracted from the image just before the progression event are correlated with the strongest absolute value to the outcome. The delta-radiomic features, so the relative changes from the follow-up MRI to the Planning MRI, and the features from the planning MRI show correlation to the label and indeed each improve the classification performance (\autoref{tab:sets_risk}), but with inferior informative power compared to the follow-up MRI.

\section{Discussion}\label{sec:discussion}

Risk classification of tumors has already been discussed in literature, although in different forms, by using radiomics to distinguish between phenotypes of tumors \cite{Aerts.2014,VanGriethuysen2017,Gillies.2016} showing different characteristics in terms of progression \cite{Lowery.2017}. 
Whereas other studies focus on improving the technique of radiomics on well known datasets \cite{Zwanenburg.2019,Zwanenburg.2020}, the emphasis in our work is to adapt the techniques to an unseen dataset. The big variety of imaging sources and inter-observer variabilites in the segmentation yield uncertainties for the performance \cite{Gillies.2016}.
Class imbalance is a typical problem in radiomics \cite{Mayerhoefer.2020} and also present in this task, as only about 5\% of the metastases show progression within the next 100 days. In a normal follow-up cycle, the metastases progressing within this interval cannot always be detected, as the imaging is only done about every 100 days. So it is to be noted that progresses, which were detected within 100 days, either were detected by coincidence or were already under observation, due to other conditions of the patient. This adds an un-accessible bias to the data.
The experiments come with further limitations. All of the work done is based on a retrospective study, meaning that all data already is acquired and put together in the dataset. Furthermore, censoring is strongly impacting the classification as already touched upon in Subsection \ref{sec:materials}. The severity of the disease leads to a limited life expectancy, 50\% of patients pass within about one year after the diagnosis of BM. Additionally, 16\% of patients left the study.

As deep learning methods take over in the analysis of images in general \cite{Litjens2017}, the next quantum leap in the quantitative analysis of medical images announces itself. Already outperforming classic radiomics in certain cases \cite{Sun2020}, the automated generation of features may improve radiomics further.

\subsection{Classification performance}

The performance of the classification is decent and comparable to further studies \cite{Gillies.2016,Aerts.2014}. The results show a good predictive value with AUC scores up to $0.83$. The limitation in number of features and the cross-validation prevent from running into the problem of overfitting \cite{Gillies.2016,Schaffer1993}. In addition, SVM as chosen algorithm for the classification is relatively unaffected by class imbalance \cite{Suthaharan2016}.
The KPM plot in \autoref{fig:kpm_split} intuitively displays the power of the classification. Hence, it tackles one of the main tasks of radiomics: Allowing communication and building bridges between computer scientists and physicians \cite{Mayerhoefer.2020}. Additionally, \autoref{fig:kpm_split} allows to assess the sensitivity and specificity graphically and not just evaluate the data at one single point in time, but for a whole timespan simultaneously.

\subsection{Impact of different feature sets}

Assessing the effect of different feature sets on the performance (See \autoref{tab:sets_risk}) leads to anticipated as well as surprising effects. As expected, the overall effect of ``more features, better results'' can be observed. Starting from inferior results when only assessing clinical data, the addition of each single feature sets improves the performance, combining these builds a broad base of information, leading to decent results. The application of wavelet filters is another leap forward, improving the performance by roughly 10\%. 
That being said, the striking effect is the influence of the Radiomic features of the planning MRI. Despite being not the latest source of information, the images of the Planning MRI already yield a lion-share of the information, used for the classification of HRM and LRM. The initial state of the BM, unadulterated by the therapy at this point in time, is already beneficial for the classification.

\subsection{Impact of single features}

In \autoref{tab:risk_all_feat}, the direct contribution of single features to the classification can be evaluated. Starting form important features from the clinical information, a lower RPA Class, lower EQD and lower number of metastasis correlate to the appearance of HRM. A lower RPA Class and number of metastasis have to be seen as ``false friends'', being more directly linked to a higher life expectation \cite{Grossman2014,Lowery.2017} and therefore keeping patients longer inside the study. A lower EQD can be very plausibly linked to the appearance of HRM. Delivering an inferior amount of dose to the tumor is insufficient to stop the progression of the tumor.

The features from the follow-up MRI state a correlation of bigger, compact lesions in the follow-up MRI to the appearance of HRM. Evaluating the wavelet-filtered images, surprisingly the features from the follow-up MRI gained more influence. All three features, originating either from the LHL or HHL transform, prefer either a high range of values or an elevated maximum intensity within the ROI, leading to pronounced yet inhomogeneous appearance of HRM candidates in these transforms.

The delta-radiomic features are defined in order to capture the changes in the Radiomic features as well as the time passed. Faster changes occurring in few days are pronounced over changes occurring in a longer timespan. This was chosen in close cooperation with physicians, to obtain a measure of fast changes regarding the metastases.
The importance of delta-radiomic features to the classification performance is inferior copared to the other Radiomic Features. Including the relative changes without any contextualization into the assessment only yields a slight benefit in the classification. However, the correlations obtained from the data are within the margins of expectation. An increase of the Delta values of “shape
-MinorAxisLength”,“shape-MajorAxisLength”, and “shape-SurfaceVolumeRatio” points to an increase in size and compactness of the metastases in short time. The progression is in the offing at the follow-up MRI and then detected at the next imaging point in time.

The features aggregated from the Planning MRI of a metastasis lead to an improvement of the classification. The information from the initial imaging is beneficial for the prediction of HRM. Interestingly, hereby the most important features also differ from the features from the follow-up MRI and are all extracted on the GLDM. The positive correlation with “gldm- LargeDependenceLowGrayLevelEmphasis” translates to and the negative correlation to the “gldm-SmallDependenceLowGrayLevelEmphasis” and “gldm-SmallDependenceEmphasis” point towards a more homogeneous structure inside the ROI. 

Although being the inferior technique for the imaging of soft tissue and brain matter \cite{Wannenmacher2013}, the PLanning CT still provides meaningful features for the classification. The higher uniformity in measurements may lead to a higher importance and better usability for Radiomic features \cite{Gillies.2016}.
The negative correlation to “shape-Sphericity” and “shape-Elongation” indicate a less sphere-shaped ROI \cite{VanGriethuysen2017}, deviating from the findings for the features of the follow-up MRI. The correlation to a lower “glszm-SizeZoneNonUniformityNormalized” translates to more homogeneous
zones of different intensity values within the ROI \cite{VanGriethuysen2017}, comparably to the GLDM-features of the Planning MRI.
The importance of the features from the Planning CT is manipulated by the choice of the ROI. The ROI within the Planning CT is defined by the PTV and therefore it does not only include tumor but also the margins around it \cite{Wannenmacher2013}. This minimizes the meaningfulness of features, describing the texture further and the assessment of the shape-features is bias, as the PTV and its margins are chosen by experts, differing from case to case \cite{Wannenmacher2013}. Thus, not solely the CT image is assessed, but also the choices by the expert, planning the treatment.

\subsection{Comparison to literature}

As already hinted at, delta-radiomics gained more and more attention in medicine over the last years \cite{Nardone2021,Fave2017}. Although the techniques and approaches differ, it has to be noted that an inclusion of different points in time allows a multifaceted assessment of the problem. As summarized by \cite{Nardone2021}, different works on tumors in the head and neck regions are based of an analysis of medical images, mostly CT images, via delta-radiomics. The classification performance is also decent with comparable or even higher AUC scores. Delta-radiomics show their great potential, on one hand in this presented work about risk classification but on the other hand also in further medical fields, far beyond the limits of head and neck cancers \cite{Nardone2021}.

\section{Summary}

The prediction of HRM that show progression within 100 days from the date of imaging, is built on a reliable dataset of medical images from the UKER. From the medical images, Radiomic features and delta-radiomic features were computed. Via the MRMR technique, the most meaningful features were selected and used to predict the outcome for the given data. Overall, over 4000 features were present per metastasis and reduced to 93, to guarantee a fast, stable, and reliable performance of the classification. The classification was realized using an SVM. The performance of the classification led to an AUC score of $0.83$ and the possibility of splitting the dataset into significantly diverging cohorts, regarding their median time to progression (LRM: $17.3$ months, HRM: $9.6$ months, $p < 0.01$). Furthermore, the importance of different features was evaluated, showing a surplus value of including multiple features from different sources: Not only the imaging right before the detection of a progression, but also the imaging at the time of the diagnosis of a BM yields information to allow a classification. The application of wavelet filters to the medical images before extracting the features showed further improvement of the classification. In more detail, higher ranges and higher maxima of the intensities in the wavelet filtered follow-up MRI, a lower RPA Class, and a lower EQD were the strongest predictive values for HRM. Overall, the findings show the capability of radiomics to predict BM with a high risk of progression and their usability on real data.

\section*{Conflict of Interest Statement}

The authors declare that the research was conducted in the absence of any commercial or financial relationships that could be construed as a potential conflict of interest.

\section*{Supplemental Data}
Code available upon time of publishing. 

\section*{Data Availability Statement}

Data are available upon request. However, legal restrictions, especially the EU General Data Protection Regulation (GDPR), the German Data Protection Laws and the Bavarian Hospital law apply, so some requests may have to be declined partially or completely.


\end{document}